%% file: conference_101719.tex
\documentclass[conference]{IEEEtran}
\IEEEoverridecommandlockouts
\usepackage{cite}
\usepackage{amsmath,amssymb,amsfonts}
\usepackage{algorithmic}
\usepackage{graphicx}
\usepackage{textcomp}
\usepackage{xcolor}
\usepackage{CJKutf8}
\def\BibTeX{{\rm B\kern-.05em{\sc i\kern-.025em b}\kern-.08em
    T\kern-.1667em\lower.7ex\hbox{E}\kern-.125emX}}
\begin{document}
\begin{CJK}{UTF8}{gbsn}
\title{Blended Latent Diffusion under Attention Control for Real-World Video Editing\\
\thanks{}
}

\author{\IEEEauthorblockN{Deyin Liu}
\IEEEauthorblockA{\textit{Department of Computer Science} \\
\textit{Swansea University}\\
Swansea, United Kingdom \\
deyin.liu@swansea.ac.uk}
~\\
\and
\IEEEauthorblockN{Lin Yuanbo Wu}
\IEEEauthorblockA{\textit{Department of Computer Science} \\
\textit{Swansea University}\\
Swansea, United Kingdom \\
l.y.wu@swansea.ac.uk}
~\\
\and
\IEEEauthorblockN{Xianghua Xie*}
\IEEEauthorblockA{\textit{Department of Computer Science} \\
\textit{Swansea University}\\
Swansea, United Kingdom \\
x.xie@swansea.ac.uk}
*Corresponding author
~\\

%\and

}

\maketitle

\begin{abstract}
Due to lack of fully publicly available text-to-video models, current video editing methods tend to build on pre-trained text-to-image generation models, however, they still face grand challenges in dealing with the local editing of video with temporal information. First, although existing methods attempt to focus on local area editing by a pre-defined mask, the preservation of the outside-area background is non-ideal due to the spatially entire generation of each frame. In addition, specially providing a mask by user is an additional costly undertaking, so an autonomous masking strategy integrated into the editing process is desirable. Last but not least, image-level pretrained model hasn't learned temporal information across frames of a video which is vital for expressing the motion and dynamics.
In this paper, we propose to adapt a image-level blended latent diffusion model to perform local video editing tasks. Specifically, we leverage DDIM inversion to acquire the latents as background latents instead of the randomly noised ones to  better preserve the background information of the input video. We further introduce an autonomous mask manufacture mechanism derived from cross-attention maps in diffusion steps. Finally, we enhance the temporal consistency across video frames by transforming the self-attention blocks of U-Net into temporal-spatial blocks. Through extensive experiments, our proposed approach demonstrates effectiveness in different real-world video editing tasks.
\end{abstract}


\section{Introduction}\label{sec:intro}
Diffusion-based generation and editing models represent a cutting-edge research area within visual content editing. Current approaches in diffusion-based editing primarily leverage large-scale pretrained text-to-image generative models,  such as Stable Diffusion \cite{Stable_Diffusion}, Imagen \cite{Imagen}, DALLE 2 \cite{DALLE_2}. 
While current methods excel in global image manipulation, there is a noticeable gap in addressing local editing needs. Local editing, which involves modifying specific regions or attributes within an image while preserving the rest, is essential for numerous practical applications, such as attribute editing and specified object manipulation. %It underscores the importance of developing techniques that can effectively handle such localized adjustments.

To enable local editing, several methods are developed: DALLE 2 \cite{DALLE_2}, GLIDE \cite{GLIDE}, Blended Diffusion \cite{Blended_diffusion}, Blended Latent Diffusion \cite{Blended_Latent_Diffusion}, FateZero \cite{FateZero}, and Video-P2P \cite{Video-P2P}. Among these methods, Blended Diffusion \cite{Blended_diffusion} stands out as a fully publicly available solution. Building upon this foundation, Blended Latent Diffusion seamlessly integrates it into the latent space of the Latent Diffusion Model \cite{Stable_Diffusion}. Subsequent advancements, including FateZero \cite{FateZero} and Video-P2P \cite{Video-P2P}, further extend and refine the capabilities of these techniques. The basic idea of Blended Latent Diffusion \cite{Blended_Latent_Diffusion} is to spatially blend the foreground latent (i.e., each of the noisy latents progressively generated in the latent denoising steps conditioned directly on the guiding text prompt) with the background latent (i.e., corresponding noised version of the original latent of the input image), by using a user-provided mask to yield the latent for the next diffusion step. However, it is problematic to employ Blended Latent Diffusion \cite{Blended_Latent_Diffusion} for local video editing due to the following reasons: \textbf{1)} For the background latent, Blended Latent Diffusion \cite{Blended_Latent_Diffusion} leverages the noised version of the original latent at the same noise level as the foreground latent. However, the added noise is stochastic, leading to inaccurate blended latent and thus affecting the local editing outcome. \textbf{2)} The blend diffusion \cite{Blended_diffusion,Blended_Latent_Diffusion} requires the user to provide a mask to specify the area to edit. This requires a user interaction or an automatic detection/segmentation method, which is of additional workload, not desirable in practice. % Why do not we consider a mask-free manner? Or, can we integrate an autonomous mask manufacture mechanism into the Blended Latent Diffusion so as to deprecate the need for specially providing the mask by another means?    
%\textbf{Thirdly}, due to the lack of sufficient text-video pairs as well as the requirement of huge computing resource for extensive training, text-to-video models are less studied in the literature. 
\textbf{3)} When one extends the original Blend Latent Diffusion method \cite{Blended_Latent_Diffusion} for video editing, it is imperative to keep the temporal consistency of the video frames, which is not learned by the pretrained text-to-image diffusion model. 

In this paper we propose to manipulate local video editing based on Blended Latent Diffusion. In specific, we first improve the preservation for the background (outside-mask area) by choosing the deterministic DDIM \cite{DDIM} inverted latents of each frame which can be used for reconstructing the input. Secondly, to avoid the need for user to provide masks, we consider an autonomous mask manufacture mechanism, which leverages the cross-attention maps \cite{p2p} from the U-Net \cite{U-net} that provides the semantic layout of the image. Such cross-attention maps can be used to produce a mask by thresholding to locate the area corresponds to the edited words. Finally, we transform the self-attention blocks in U-Net \cite{U-net} into temporal-spatial attention blocks to capture the inter-frame temporal information and enhance the temporal consistency of video appearance.

Our proposed video editing method termed Blend Latent Diffusion under Attention Control consists of the above three modules: DDIM inversion based latents for background, cross-attention control for automatic masking and temporal-spatial attention for temporal consistency enhancement. We conduct extensive experiments to verify the superior performance of our method by comparing with the state-of-the-art video editing methods and ablation studies. %It is worth noting that though existing methods FateZero \cite{FateZero} and Video-P2P \cite{Video-P2P} use DDIM inversion, FateZero \cite{FateZero} store DDIM inversion and then blend the attention maps corresponding to the source and target prompts, while Video-P2P \cite{Video-P2P} tunes the model to acquire an optimized video inversion based on DDIM inversion. Tune-A-Video \cite{TAV} also considers to adapt the self-attention to the spatial-temporal self-attention to describe the temporal consistency in the video. Nonetheless, Tune-A-Video \cite{TAV} needs to update the weights of attention blocks. 

\begin{figure*}
\centering
\includegraphics[width=\linewidth]{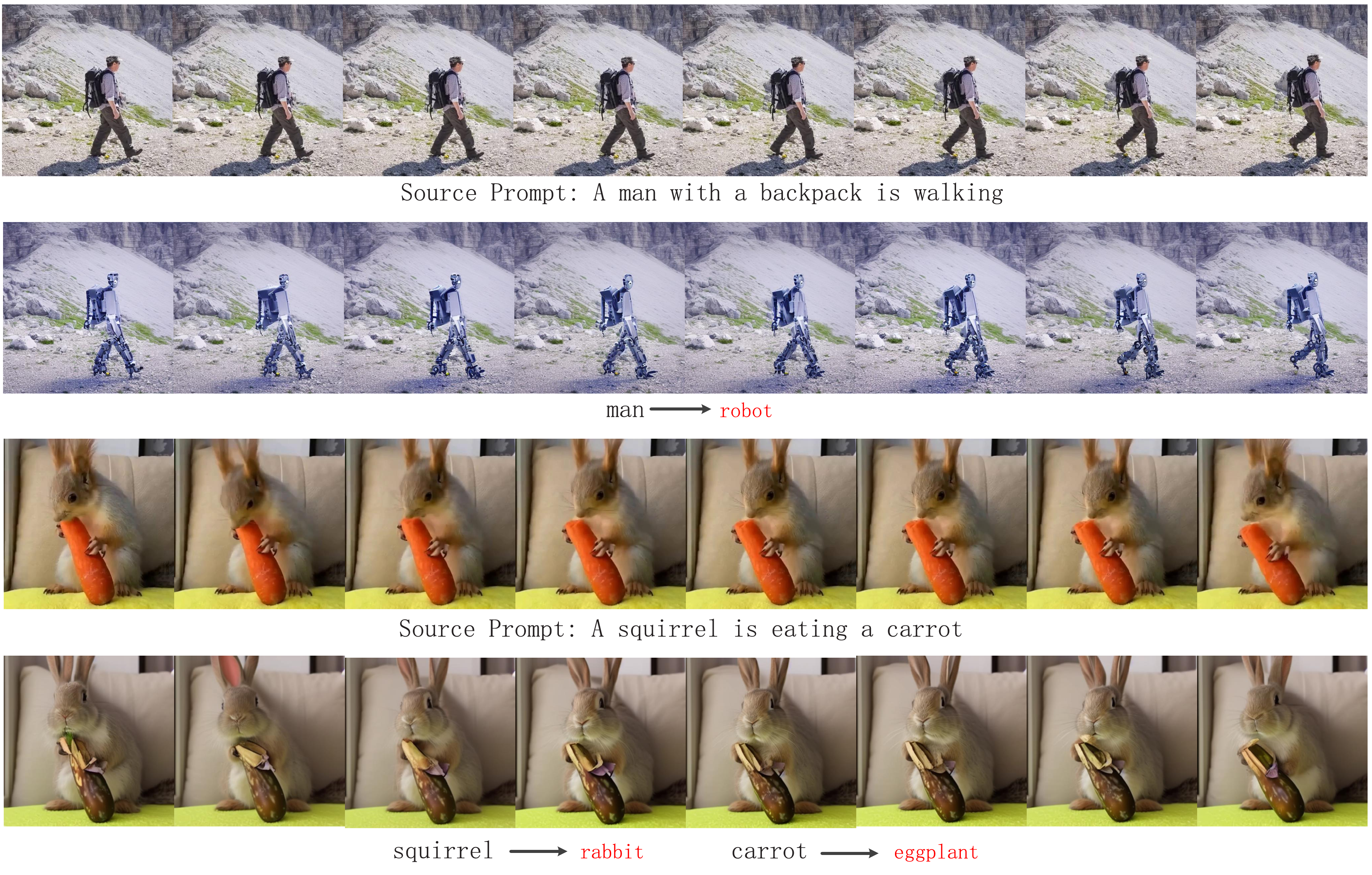}
\caption{Examples of local video editing achieved by our proposed method}
\label{fig1}
\end{figure*}

\input{related}

\section{The Method}\label{sec:method}
\subsection{Preliminary}
\textit{Latent Diffusion Models}\cite{Blended_diffusion} (e.g., Stable Diffusion) extend the diffusion models \cite{DDPM} into the latent space of an autoencoder. Firstly, an encoder $E$ compresses an image $x$ to a lower dimensional latent $z = E(x)$, which can be reconstructed back to image $D(z) \approx x$ via a decoder $D$. Thereafter, a U-Net \cite{U-net} $\epsilon_\theta$ consisting of cross-attention and self-attention blocks is trained to predict the artificial noise using the following objective:
\begin{equation}\label{eqn-1}
\min\limits_\theta E_{z_0,\varepsilon\sim N(0,I), t\sim \text{Uniform}(1,T)}\left\lVert\varepsilon - \varepsilon_\theta (z_t, t, p) \right\rVert_2^2,
\end{equation}
where $p$ is the embedding of the conditional text prompt and $z_t$ is a noisy sample of $z_0$ at timestep $t$.

\textit{Denoising Diffusion Implicit Models (DDIM)}\cite{DDIM} is a deterministic sampling technique, which is employed during inference to convert a random noise $z_T$ to a clean latent $z_0$ in a sequence of timestep $t: T \longrightarrow 1$:
\begin{equation}\label{eqn-2}
z_{t-1}=\sqrt{\alpha_{t-1}}\frac{z_t - \sqrt{1-\alpha_t}\varepsilon_\theta}{\sqrt{\alpha_t}}+ \sqrt{1-\alpha_{t-1}}\varepsilon_\theta,
\end{equation}
where $\alpha_t$ is a parameter for noise scheduling and each $\varepsilon_\theta$ stands for the noise $\varepsilon_\theta (z_t, t, p)$  predicted at timestep $t$.

\textit{Blended Latent Diffusion}\cite{Blended_Latent_Diffusion} follows the idea of Blended Diffusion \cite{Blended_diffusion} and repeatedly blends two parts, i.e., foreground and background, into the latent space as the diffusion progresses. The foreground (fg) refers to the part that one wishes to modify, which is to be restricted by a given mask $m$, while the background (bg) refers to the rest. Specifically, in the latent space, due to the convolutional nature of the autoencoder, the width and the height are smaller than those of the input image (by a scalability factor of 8). The input mask $m$ is therefore downsampled to such spatial dimensions to obtain the latent space binary mask $m_{latent}$, which will be used to perform the blending. Then, the denoising diffusion process is manipulated in the following way: at each step, a latent denoising step is first performed, conditioned directly on the embedding of the guiding text prompt $p_{edit}$, to obtain a less noisy foreground latent denoted as $z_{fg}$. Meanwhile, the original latent $z_0$ is noised and injected into the current noise level in a step (via $\sqrt{\overline{\alpha}_t}z_0 + \sqrt{1-\overline{\alpha}_t}\varepsilon$) to obtain a noisy background latent $z_{bg}$. The two latents $z_{bg}$ and $z_{fg}$ are then blended using the resized mask to yield the latent for the next latent diffusion step via:
\begin{equation}\label{eqn-3}
z_{t-1}=z_{fg}\odot m_{latent}+z_{bg}\odot (1-m_{latent}),
\end{equation}
where $\odot$ is element-wise multiplication. At each denoising step the latent is modified corresponding to the edit prompt, whilst the subsequent blending enforces the part outside $m_{latent}$ to remain the same. Though the resulting blended latent is not guaranteed to be coherent between foreground and background, the next latent denoising step can address it. Once the latent diffusion process terminates, the final latent is decoded to the output image using the decoder $D(z)$.

\subsection{Blended Latent Diffusion under Attention Control}

\textbf{\textit{DDIM Inversion for Background Latents}}: As Blended Latent Diffusion \cite{Blended_Latent_Diffusion} suggested, the background latent $z_{bg}$ is chosen as the noised version of the initial latent of the image at the same noise level as the foreground latent, although the foreground and the background latents can be viewed as riding on the same manifold, the added noise for the background latent is random, not deterministic, thus the original intention of preserving the outside-mask area in the image is not well achieved. In contrast, DDIM inverted latents, which can be used  to progressively reconstruct the original latent, will be a better alternative for the background. According to the ODE limit analysis of the diffusion process, DDIM inversion\cite{DBeatG,DDIM} is able to map the initial latent $z_0$ to a sequence of noised latents $z_t$ in the steps $t : 1 \longrightarrow T$, which is reverse to DDIM sampling in Equation \eqref{eqn-2}:
\begin{equation}\label{eqn-4}
\hat{z}_{t}=\sqrt{\alpha_{t}}\frac{\hat{z}_{t-1} - \sqrt{1-\alpha_{t-1}}\varepsilon_\theta}{\sqrt{\alpha_{t-1}}}+ \sqrt{1-\alpha_{t}}\varepsilon_\theta.
\end{equation}
In such a way, the acquired latents $z_t$, $t:1 \longrightarrow T$, can be used to accurately reconstruct the initial latent. Thus, we can choose them as the background $z_{bg}$ to replace the randomly noised latents to preserve the background area.

\textbf{\textit{Cross Attention Thresholding Mask}}: Unlike Blended Latent Diffusion \cite{Blended_Latent_Diffusion} that needs a user-provided mask to localize the edit to the specified area, we take advantage of the cross-attention map which provides the semantic layout of the image closely related to the prompt text \cite{p2p}. Hence, we can obtain a binary mask $m_t$ by thresholding the cross-attention maps of the edited words during the diffusion process by a constant $\tau$. Specifically, we first compute the average attention map $\overline{M}_{t,w}$ (averaged over steps $T, T-1, \ldots, t$) of the source word $w$ during the DDIM inversion conditioned on the embedding of the source text prompt $p_{src}$, and then we calculate the average attention map $\overline{M}^{\ast}_{t,w^{\ast}}$ of the new word $w^{\ast}$ during the diffusion process conditioned on the embedding of the target text prompt $p_{edit}$. A threshold is set to produce the binary maps, where $B(x) := x >\tau$ and $\tau= 0.3$ throughout all the experiments. To achieve seamless local editing, the edited region should include the silhouettes of both the source and the newly edited area. To this end, the final mask $m_{latent}$ is a union of the two binary maps:
\begin{equation}\label{eqn-5}
m_{latent} = B(\overline{M}_{t,w}) \cup B(\overline{M}^{\ast}_{t,w^{\ast}}).
\end{equation}
Using such an autonomous mask manufacture mechanism can avoid the cumbersome provision of a mask from a pre-defined method such as image segmentation.

\textbf{\textit{Temporal-Spatial Attention}}: The previous two designs construct a strong local editing method that can preserve the structure of source image with the background almost retained.  When applied to video editing, however, denoising each frame individually readily produces temporally inconsistent video since the pretrained text-to-image model used cannot learn the temporal information regarding inter-frame consistency. Inspired by the recent one-shot video generation method \cite{TAV}, we reshape the original self-attention to temporal-spatial attention with the pretrained weights unchanged. Specifically, for a video consisting of $n$ frames, we implement the $\text{ATTENTION}(Q, K, V)$ \cite{transformer} for the original self attention input feature $z_i$ at frame index $i \in [1, n]$ as:
\begin{equation}\label{eqn-6}
Q=W^Q z^i, K=W^K[z^{i-1};z^i], V=W^V[z^{i-1};z^i],
\end{equation}
where $[\cdot]$ denotes the concatenation operation. $W^Q$, $W^K$, $W^V$ are the projection matrices from the pretrained model. 
Then, the temporal-spatial attention map is represented as $s_t \in R^{hw*fwh}$, where $f$=2 is the number of frames used as key and value. It captures both the structure of a single frame and the temporal correspondence with the nearest neighbor frame features. In such a training-free way, we build up the relationships of each frame with its nearest neighbor, maintaining the temporal continuity and inter-frame consistency.

\section{Experiments}\label{sec:exp}
\subsection{Implementation Details}
We choose the pretrained CompVis stable diffusion v1.4 \cite{Stable_Diffusion} as the base model, and the DDIM \cite{DDIM} sampling and inversion are generally with total timestep $T = 50$. The threshold for the binary mask is set to $\tau= 0.3$. For evaluation, we use the videos from DAVIS \cite{DAVIS}, Edinburgh office monitoring video dataset \cite{Edinburgh} and other in-the-wild videos, and generally we sample 8 frames from each video for use. The source prompts of the videos are generated via the image caption model \cite{BLIP} while the target prompt for each video is designed mainly by replacing several words.

\subsection{Applications}

Based on the pretrained text-to-image diffusion model \cite{Stable_Diffusion}, our proposed method can be used for local attribute editing, local object property editing, or local object category replacement in a video as shown in Fig.1, Fig.2 and Fig.3. 
In the first and second rows of Fig.1, the proposed method changes the whole appearance of an original object, the hiker, to a robotic one with the same dynamics of walking, by just replacing the word "man" in the source prompt into "robot" in the target prompt. Similarly, in Fig.2, the attribute of the floor of the playground is changed from "cement floor" into "ice surface" corresponding to the change in the prompts. In Fig.3, the "swan" is changed into "duck" with local modifications especially the shape of the beak. And in the last two rows of Fig.1, the video is edited from "A squirrel is eating a carrot" to "A rabbit is eating an eggplant", completely replacing two local objects' categories into another two. When editing the categories of objects, it is challenging because of the thorough change of shapes and appearances, with the action, pose and position similar to the input video, keeping the motion or temporal dynamics unchanged.

\begin{table}[t]
\caption{Quantitative evaluation for different methods}
\begin{center}
\centering
\begin{tabular}{|p{0.4\columnwidth}|p{0.2\columnwidth}|p{0.2\columnwidth}|}
\hline
\textbf{Inversion and Editing }&\multicolumn{2}{|c|}{\textbf{CLIP Metrics}} \\
\cline{2-3} 
\textbf{Methods} & \textbf{\textit{Frame-wise Accuracy}}& \textbf{\textit{Temporal Consistency}} \\
\hline
Frame-wise Null-text inversion and Prompt-to-prompt \cite{p2p,NTI} & 0.96 & 0.85   \\
\hline
Frame-wise SDEdit \cite{SDEdit} & 0.82  & 0.91   \\
\hline
NLA, Null-text inversion and Prompt-to-prompt \cite{LNA,NTI,p2p} & 0.60 &  0.95 \\
\hline
DDIM inversion and Tune-A-Video \cite{TAV,DDIM} & 0.75 &   0.96  \\
\hline
FateZero \cite{FateZero} & 0.90 & 0.97 \\
\hline
The proposed  & 0.91 & 0.95 \\
\hline
\end{tabular}
\label{tab1}
\end{center}
\end{table}

\subsection{Comparisons with State-of-the-Arts}
We compare our method with the following state-of-the-art methods: (1) Frame-wise Null-text optimization \cite{NTI} combined with prompt-to-prompt \cite{p2p}; (2) Frame-wise editing method SDEdit \cite{SDEdit}; (3) Tune-A-Video \cite{TAV}; (4) The Neural Layered Atlas (NLA) \cite{LNA} based method combined with key frame editing \cite{p2p,NTI}; (5) Fusing Attentions for Zero-shot Text-based Video Editing (FateZero) \cite{FateZero}. For attention-based editing methods, the  timestep parameters are set to identical to us.

In our experiments, quantitative evaluation are also conducted utilizing the trained CLIP \cite{CLIP} model, following existing methods \cite{Patrick,I2I,TAV}. The results are shown in Table 1. The ``Frame-wise Accuracy'' \cite{CLIPScore} is the frame-wise editing accuracy and it represents the percentage of frames for which the edited image has a higher CLIP score to the target prompt than to the source prompt. The ``Temporal Consistency'' \cite{Patrick} measures the temporal consistency among frames, which is an average of cosine similarities of
all pairs of consecutive frames. 
Table 1 shows that the proposed method achieves comparable
frame-wise editing accuracy and temporal consistency as the state-of-the-art methods in local video editing tasks. 
Although Frame-wise null inversion \cite{NTI} achieves the best Frame-wise Accuracy, it costs 500 iterations of optimization for each frame, with low temporal consistency. It's known that NLA-based method \cite{LNA} needs to take hours to optimize the neural atlas for each input video. Tune-A-Video \cite{TAV} combined with DDIM inversion and FateZero \cite{FateZero} show impressive Temporal Consistency, which benefits from the finetune for spatial-temporal self attention and the temporal attention modules, especially when applied in shape-aware editing. In contrast, our method just reshapes the self attention into temporal-spatial attention with all the weights unchanged, achieving a comparably superior temporal consistency in a training-free way.

\subsection{Ablation Studies}

In this section, we ablate our method at different components in the video editing.
Firstly, we study the effect of DDIM inversion for background latents. In the blended latent diffusion, for the choice of the background latents, we first use the previous setting, the randomly noised version of the original latent at the same noise level of the each foreground latent, and then the performance of one edited video example is shown in the third row of Fig.2. The target is to edit a local attribute of the original video, changing the cement floor to an ice surface. Although the target is achieved basically in the third row, as you can see by zooming in, some details in the background area (here it refers to other part other than the floor in the video) are changed too, for example, the face of the player has been "worn" a face shield, and his shoes are changed from inline skates to blade ice skates.  
In contrast, as shown in the second row, we use DDIM inverted latents instead of the randomly noised to overcome these problems, keeping the background area unchanged even in such details. This indicates that the latents resulted from DDIM inversion are better choice to preserve the background information. 

\begin{figure}
\centering
\includegraphics[width=\columnwidth]{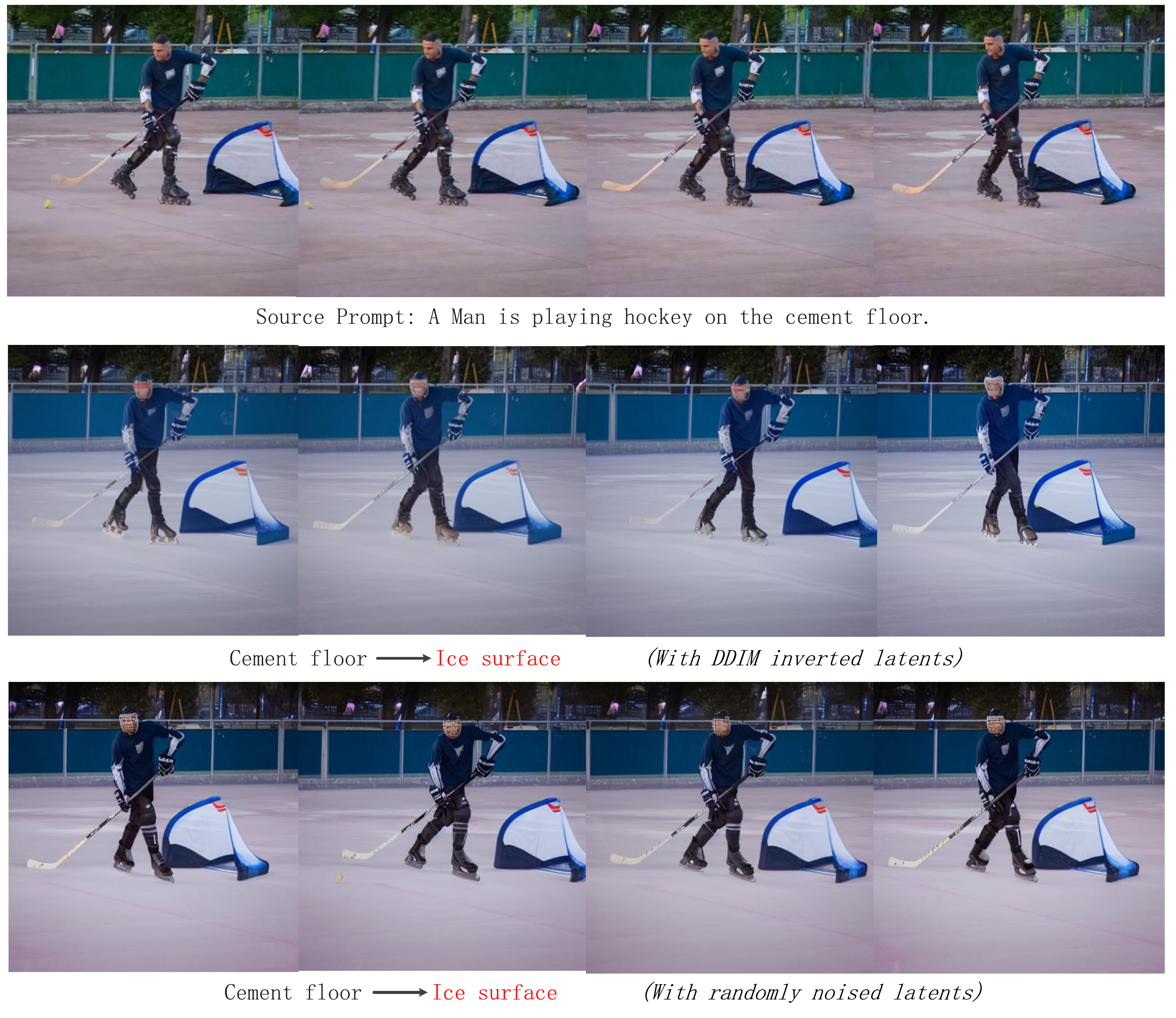}
\caption{Performance comparisons between using the DDIM inverted latents and previous randomly noised ones}
\label{ablation_fig2}
\end{figure}
% \begin{figure}[htbp]
% \centerline{\includegraphics{ablation_fig1}}
% \caption{Example of Ablation Fig1.}
% \label{fig}
% \end{figure}

\begin{table}[t]
\caption{Quantitative comparisons regarding the effect of mask}
\begin{center}
\centering
\begin{tabular}{|p{0.4\columnwidth}|p{0.2\columnwidth}|p{0.2\columnwidth}|}
\hline
\textbf{Mask }&\multicolumn{2}{|c|}{\textbf{CLIP Metrics}} \\
\cline{2-3} 
\textbf{Schemes} & \textbf{\textit{Frame-wise Accuracy}}& \textbf{\textit{Temporal Consistency}} \\
\hline
Without Mask    & 0.55 & 0.81 \\
\hline
User-provided Mask (with the dataset)    & 0.92 & 0.95 \\
\hline
Cross Attention Thresholding Mask    & 0.91 & 0.95 \\
\hline
\end{tabular}
\label{tab2}
\end{center}
\end{table}

Secondly, we evaluate the effect of the mask in the local video editing. The quantitative comparisons among different cases are shown in Table 2. Obviously, the one without any mask has the worst local editing performance since there is no any means to limit the target edit into a local area. We can also find that our proposed autonomous mask manufacture scheme, Cross Attention Thresholding Mask presents almost the same local editing performance as the User-provided Mask, while the autonomous scheme will save a lot of extra workload in providing such masks.

\begin{figure}
\centering
\includegraphics[width=\columnwidth]{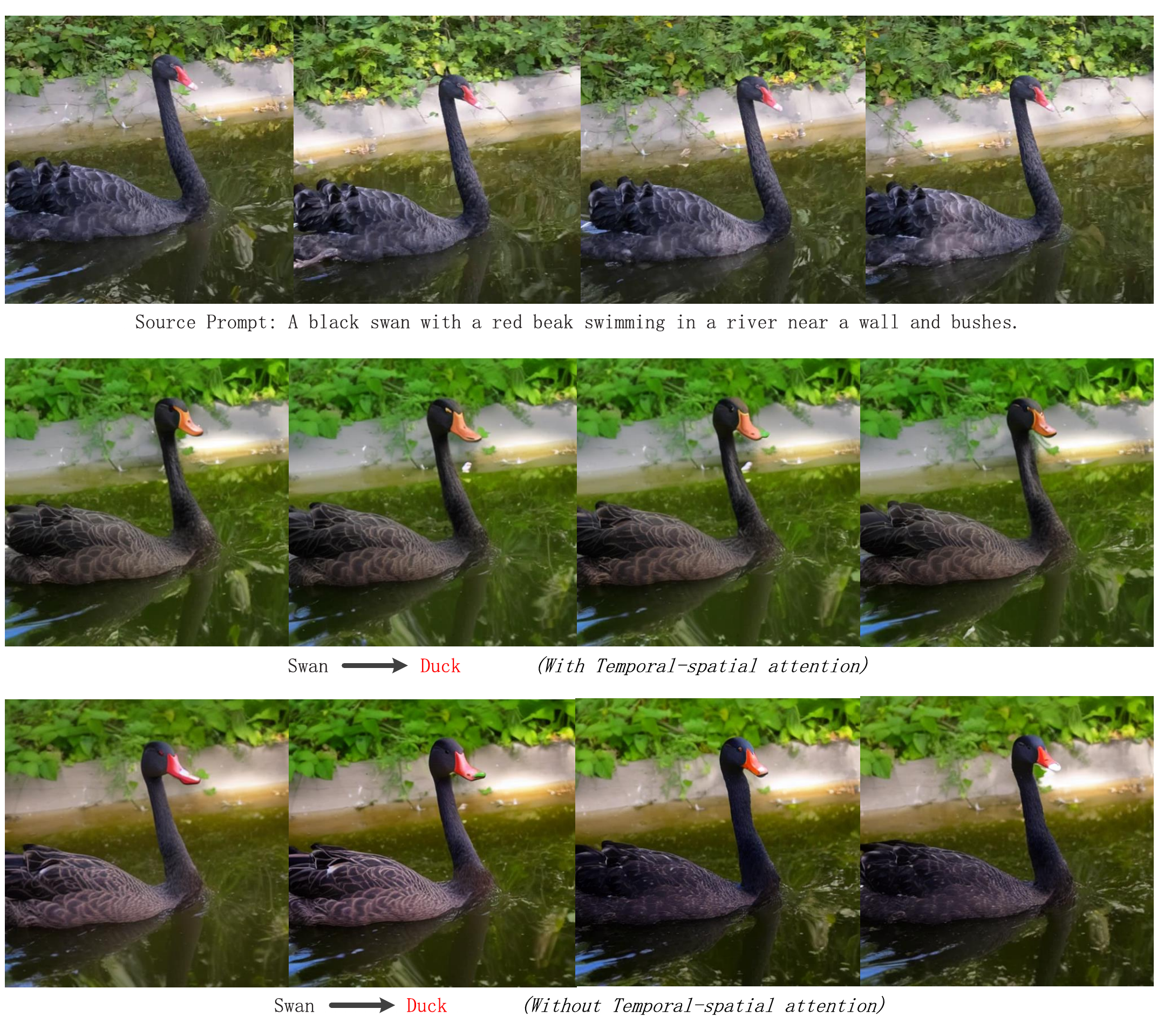}
\caption{Performance comparisons with/without the proposed temporal-spatial attention}
\label{ablation_fig3}
\end{figure}

Thirdly, we analyze the role of the Temporal-spatial attention module. As shown in Fig.3, the performance of the second row is with Temporal-spatial attention, while the third row is without it. Through careful observation, it is not hard to find that, in the third row, the shape and color layout (red part and white point) of the edited duck's beak are very inconsistent among the four frames, and the color tones of the duck are also different (the former two frames are light black, while the last two are heavey black). In contrast, the second row shows all consistent in these details, illustrating the significance of the Temporal-spatial attention in keeping the inter-frame consistency.

\section{Conclusion}
In this paper, we propose a novel text-driven video editing framework that performs local video object editing subject to user-prompts. We utilise the DDIM inverted latents to serve as background and blends with the foreground latents during denoising steps. To accurately localise the local area, we develop an automatic masking mechanism leveraging the cross-attention maps corresponding to the edited words. We further transform the self-attention block in U-Net architecture into temporal-spatial attention, which enhances the temporal consistency in the video. We also demonstrate the proposed method's impressive effectiveness in various local video editing tasks on attribute change,  object replacement and so on.

\section*{Acknowledgment}

This project was funded by Airbus Endeavr Wales.

\bibliography{allbib}
\bibliographystyle{IEEEtran}
\end{CJK}
\end{document}

%% file: related.tex
\section{Related Work}\label{sec:related}

\subsection{Text-to-Image Synthesis}

Text-to-image synthesis has garnered increasing interest in recent years, which generates images that match the given textual description in terms of both semantic consistency and image realism \cite{LIU-TNNLS-Verbal-PersonNet}. Seminal works based on RNNs \cite{Generate-RNN} and GANs \cite{AttnGAN,StackGAN,StackGAN-pp}, were later superseded by transformer-based approaches \cite{Attention-all-need}. For example, DALL·E \cite{DALLE} proposed a two-stage approach where in they first trained a discrete VAE \cite{Discrete-VAE} to learn a rich semantic context, and then they trained a transformer model to autoregressively model the joint distribution over the text and image tokens.

Diffusion models were also used for various global image-editing applications. ILVR \cite{ILVR} demonstrates how to condition a DDPM model \cite{DDPM} on an input image for image translation tasks. Palette \cite{Palette} trains a designated diffusion model to perform four image-to-image translation tasks, namely colorization, inpainting, uncropping, and JPEG restoration. SDEdit \cite{SDEdit} demonstrates stroke painting to image, image compositing, and stroke-based editing. RePaint \cite{RePaint} uses a diffusion model for free-form inpainting of images.

\subsection{Text Driven Video Editing}

It becomes more challenging to edit the object shape in real-world videos. Current methods exhibit artifacts even with the optimization on generative priors \cite{Shape-aware}. The stronger prior of the diffusion-based model also draws the attention from researchers, e.g., gen1 [9] trains a conditional model for depth and text-guided video generation, which can edit the appearance of the generated images on the fly. Dreamix [36] finetunes a stronger diffusion-based video model [18] for editing with stronger generative priors. Nonetheless, both of these methods need private and powerful video diffusion models for editing.